\documentclass[conference]{IEEEtran}
\usepackage{cite}
\usepackage{amsmath,amssymb,amsfonts}
\usepackage{algorithmic}
\usepackage{graphicx}
\usepackage{textcomp}
\usepackage{xcolor}
\usepackage[frozencache,cachedir=.]{minted}
\usepackage{multirow}
\usepackage[T1]{fontenc}
\usepackage{enumitem}
\usepackage{hyperref}
\usepackage{fancyhdr}

\newcommand{\rev}[1]{{{#1}}}

\def\BibTeX{{\rm B\kern-.05em{\sc i\kern-.025em b}\kern-.08em
    T\kern-.1667em\lower.7ex\hbox{E}\kern-.125emX}}

\begin{document}

\title{Sparse GPU Kernels for Deep Learning}

\author{
\IEEEauthorblockN{Trevor Gale}
\IEEEauthorblockA{
\textit{Stanford University}\\
United States of America \\
tgale@cs.stanford.edu}
\and
\IEEEauthorblockN{Matei Zaharia}
\IEEEauthorblockA{
\textit{Stanford University}\\
United States of America \\
matei@cs.stanford.edu}
\and
\IEEEauthorblockN{Cliff Young}
\IEEEauthorblockA{
\textit{Google Brain}\\
United States of America \\
cliffy@google.com}
\and
\IEEEauthorblockN{Erich Elsen}
\IEEEauthorblockA{
\textit{DeepMind}\\
United Kingdom \\
eriche@google.com}}

\maketitle
\thispagestyle{fancy}
\lhead{}
\rhead{}
\chead{}
\lfoot{\footnotesize{
SC20, November 9-19, 2020, Is Everywhere We Are
\newline 978-1-7281-9998-6/20/\$31.00 \copyright 2020 IEEE}}
\rfoot{}
\cfoot{}
\renewcommand{\headrulewidth}{0pt}
\renewcommand{\footrulewidth}{0pt}

\begin{abstract}
Scientific workloads have traditionally exploited high levels of sparsity to accelerate computation and reduce memory requirements. While deep neural networks can be made sparse, achieving practical speedups on GPUs is difficult because these applications have relatively moderate levels of sparsity that are not sufficient for existing sparse kernels to outperform their dense counterparts. In this work, we study sparse matrices from deep learning applications and identify favorable properties that can be exploited to accelerate computation. Based on these insights, we develop high-performance GPU kernels for two sparse matrix operations widely applicable in neural networks: sparse matrix--dense matrix multiplication and sampled dense--dense matrix multiplication. Our kernels reach 27\% of single-precision peak on Nvidia V100 GPUs. Using our kernels, we demonstrate sparse Transformer and MobileNet models that achieve \rev{1.2--2.1$\times$} speedups and up to 12.8$\times$ memory savings without sacrificing accuracy.
\end{abstract}


\setlength{\textfloatsep}{1mm}
\setlength{\floatsep}{1mm}
\setlength{\dbltextfloatsep}{1mm}
\setlength{\abovecaptionskip}{1mm}

\section{Introduction}
Deep neural network architectures are composed of large, dense matrices used in matrix multiplication and convolutions \cite{resnet, transformer}. These matrices can be made sparse with little to no loss in model quality, leading to models that are more efficient in terms of both the floating-point operations (FLOPs) and parameters required to achieve a given accuracy \cite{lwac, exploring-sparsity, to-prune-or-not, sparse-transformer}.

\begin{figure}[ht]
  \includegraphics[width=\columnwidth]{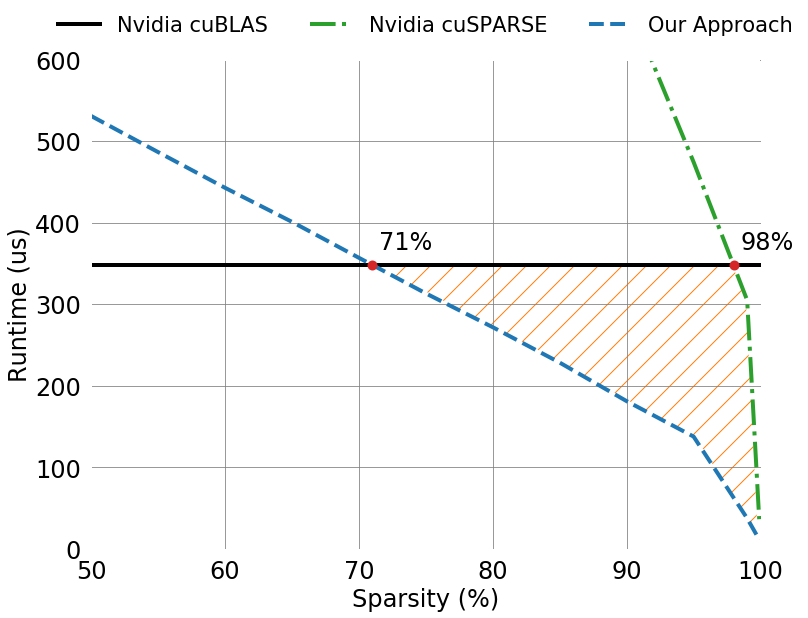}
  \vspace{-6mm}
  \caption{\textbf{Sparse matrix--matrix multiplication runtime for a weight-sparse long short-term memory network problem}. Input size 8192, hidden size 2048, and batch size 128 in single-precision on an Nvidia V100 GPU with CUDA 10.1. Using our approach, sparse computation exceeds the performance of dense at as low as 71\% sparsity. Existing vendor libraries require 14$\times$ fewer non-zeros to achieve the same performance. This work enables speedups for all problems in the highlighted region.}
\label{spmm_gemm_crossover}
\end{figure}

\begin{figure*}[ht!]
  \includegraphics[width=\textwidth]{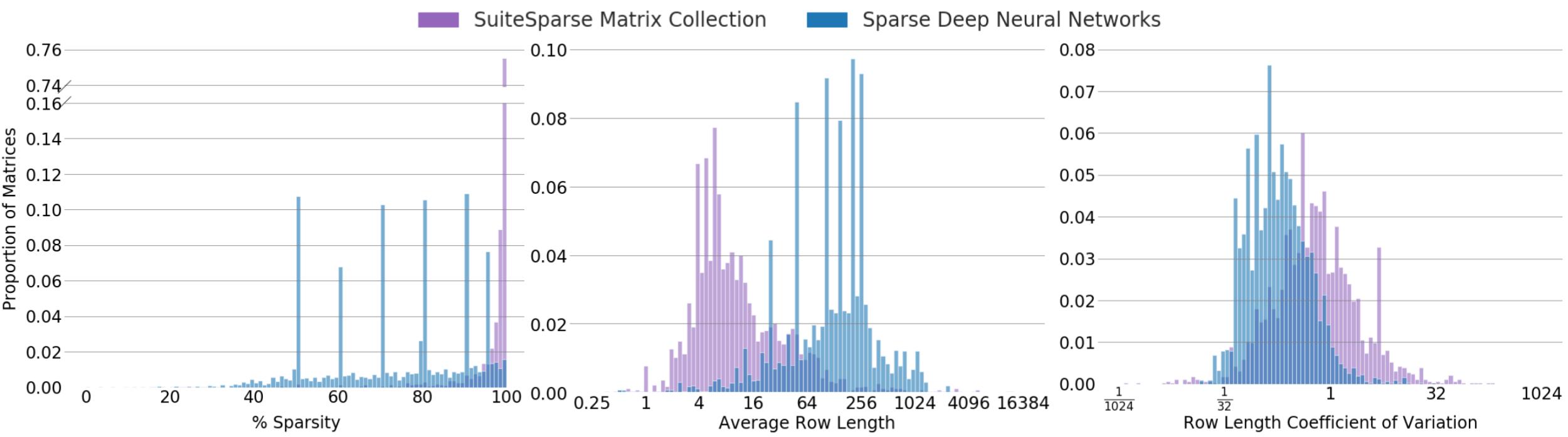}
  \vspace{-6mm}
  \caption{\textbf{Properties of sparse matrices from scientific computing and deep learning applications.} Histograms are partially transparent to show overlapping regions. On average, deep learning matrices are 13.4$\times$ less sparse, have 2.3$\times$ longer rows, and have 25$\times$ less variation in row length within a matrix.
}
\label{matrix_study}
\end{figure*}

The most common use of sparsity in deep neural networks is to accelerate inference. In addition to the standard training procedure, a sparsification algorithm is applied to produce a neural network where a high fraction of the weights are zero-valued \cite{optimal-brain-damage, lwac, sparse-variational-dropout, l0-regularization}. The weight matrices can then be stored in a compressed format, and sparse linear algebra kernels can be used to accelerate computation. In the context of generative models, sparsity has been applied to reduce the computational requirements of self-attention in Transformer architectures \cite{sparse-transformer, reformer, routing-transformer}. In addition to these applications, sparsity can be exploited to achieve higher predictive accuracy by training a larger, sparse model for a fixed computational cost \cite{blocksparse-rnn, blocksparse-gpu-kernels, wavernn}. To make training large sparse models feasible, all computation during training needs to operate directly on the compressed sparse representation of the model's weights.

The potential applications of sparsity in deep learning are numerous. However, it is difficult to realize the benefits of sparsity in real applications due to the lack of efficient kernels for core sparse matrix computations like sparse matrix--matrix multiplication (SpMM) and sampled dense--dense matrix multiplication (SDDMM) on accelerators like GPUs. 

On parallel architectures, the performance of sparse linear algebra kernels can vary drastically with properties of the sparse matrix such as the topology of nonzero values and level of sparsity. Existing GPU kernels for sparse linear algebra are primarily optimized for scientific applications, where matrices are extremely (99\%+) sparse. With the relatively moderate levels of sparsity found in deep neural networks, these kernels are not able to outperform their dense counterparts.

To address this issue, structure can be enforced on the topology of nonzeros such that nonzero values are grouped into blocks \cite{blocksparse-rnn, blocksparse-gpu-kernels, wavernn}. While this approach is able to recover much of the performance achieved by dense computation, the constraint on the location of nonzeros can significantly degrade model quality relative to unstructured sparsity \cite{wavernn, balanced-sparsity, fast-sparse-convnets}.

In this work, we develop an approach for computing SpMM and SDDMM on GPUs which is targeted specifically at deep learning applications. Our approach operates directly on the standard compressed sparse row (CSR) format and does not enforce any structure on the topology of nonzero values. \textbf{We make the following specific contributions:}
\begin{itemize}
\item We conduct a large-scale study of sparse matrices found in deep learning and identify favorable properties that can be exploited to accelerate sparse computation.
\item We introduce a \textit{1-dimensional tiling} scheme for decomposing the computation across processing elements that facilitates reuse of operands and lends itself to an extensible implementation.
\item We develop two techniques, \textit{subwarp tiling} and \textit{reverse-offset memory alignment}, that enable the use of vector memory instructions on misaligned memory addresses in sparse data structures.
\item We introduce \textit{row swizzle load balancing}, an approach for load balancing computation between processing elements that is decoupled from the parallelization scheme.
\end{itemize}


On a large dataset of sparse matrices taken from state-of-the-art deep neural networks, we demonstrate geometric mean speedups of 3.58$\times$ and \rev{2.19$\times$} over Nvidia cuSPARSE for SpMM and SDDMM respectively on Nvidia V100 GPUs. On the top performing problems, our  kernels  reach  27\%  of  single-precision  peak. Using our kernels, we demonstrate sparse Transformer and MobileNet models that achieve \rev{1.2--2.1$\times$} end-to-end speedups and 12.8$\times$ reductions in memory usage while matching the accuracy of their dense counterparts. Our code is open-source and available at \url{https://github.com/google-research/sputnik}


\section{Sparse Matrices in Deep Learning}

To understand the properties of sparse matrices in deep learning, we constructed a dataset of sparse deep neural network weight matrices from the large-scale study of \cite{tsos}. The dataset is composed of ResNet-50 \cite{resnet} and Transformer \cite{transformer} models trained on ImageNet \cite{imagenet} and WMT14 English-to-German \cite{wmt14} respectively, and includes models trained with four different algorithms for inducing sparsity in neural networks. For Transformer, we limit our analysis to models that achieve above 20 BLEU on the WMT14 English-German test set. For ResNet-50, we include models that achieve over 70\% top-1 accuracy on the ImageNet validation set. In total, the collection includes 3,012 matrices from 49 different models.

Our analysis focuses on three properties of the matrices: row length (in number of nonzeros) coefficient of variation (CoV), average row length, and sparsity. The CoV of a matrix's row lengths is the standard deviation of the row lengths divided by their mean. A high CoV is indicative of load imbalance across the rows of a sparse matrix. The average row length captures the average amount of work that will be done on each row of the sparse matrix. Longer row lengths are desirable as startup overhead and one-time costs can be amortized over more useful work. Sparsity measures the fraction of values that are zero valued in a matrix. Depending on the implementation, lower sparsity levels can be useful to increase the likelihood that nonzero values in different rows fall into the same columns, opening up opportunities for the reuse of operands through caches.

We contrast the properties of deep learning workloads with matrices from the SuiteSparse Matrix Collection \cite{suitesparse}, which is made up of 2,833 sparse matrices from a wide range of scientific workloads including circuit simulations, computational fluid dynamics, quantum chemistry, and more.

\begin{figure*}[t]
  \includegraphics[width=\textwidth]{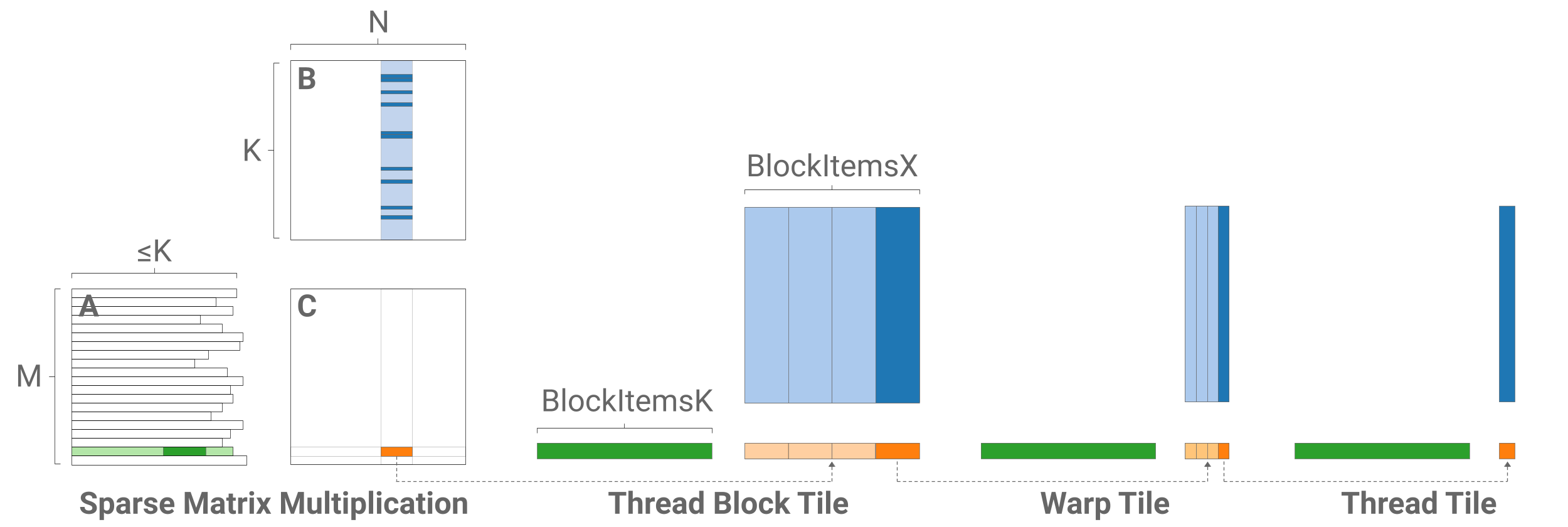}
  \vspace{-6mm}
  \caption{\textbf{Hierarchical decomposition of SpMM with 1-dimensional tiling}. Visualized with 4 warps in a thread block and 4 threads in a warp for brevity. In each level of the decomposition, matrix $A$ is a sparse matrix stored in compressed-sparse row format, marked in green and shown on the left. Matrix $B$ is dense, marked in blue and shown on top of the output matrix. The output matrix $C$ is dense, marked in orange and shown in the bottom right of each level. The dark regions at each level denote data used by the following level. \textbf{Far Left:} Each thread block computes a 1-dimensional tile of the output matrix. All values from the row of sparse matrix $A$ are needed by all threads. We use all threads in the thread block to collaboratively load values and indices from $A$ to shared memory where they can be quickly accessed for computation. For each column index loaded from $A$, the thread block will load a contiguous vector from matrix $B$ (marked with dark blue horizontal stripes). \textbf{Center Left - Far Right:} Threads process independent outputs and thus need disjoint subsets of columns from dense matrix $B$. Each thread loads the values from $B$ needed to compute it's outputs and stores them in thread-local registers.
}
\label{tiling}
\end{figure*}

\subsection{Results \& Analysis}

Statistics for our corpus of deep learning matrices and the SuiteSparse Matrix Collection are plotted in Figure \ref{matrix_study}. The difference between sparse matrices from scientific workloads and those from deep learning is considerable: on average, deep learning matrices are 13.4$\times$ less sparse, have 2.3$\times$ longer rows, and have 25$\times$ less variation in row length within a matrix. We find it likely that these differences are primarily driven by the desire to maintain high accuracy, which requires deep neural networks with a large number of parameters. This in turn leads to a higher number of nonzeros per row and a lower CoV, which is inversely proportional to average row length. For each of the metrics that we studied, deep learning matrices exhibit favorable properties that we can take advantage of to accelerate sparse matrix computations.

\section{Graphics Processing Units Background}
This section provides a basic description GPU architecture and terminology. Our implementation is written in CUDA and thus we opt for the terminology used by Nvidia. 

GPUs are made up of an array of \textit{streaming multiprocessors} (SMs) and GPU kernels are made up of threads that are grouped into sets of 32 called \textit{warps}. Warps are grouped into larger sets of threads called \textit{thread blocks}. The set of thread blocks that make up a kernel is called a \textit{grid}. When a kernel is launched to the GPU for execution, each thread block is scheduled onto an SM. A \textit{wave} of thread blocks is a set of thread blocks that run concurrently on the GPU \cite{gpu-perf}. All threads within a thread block can communicate through fast, programmer-managed, shared memory that is local to the SM. All threads also have access to thread-local registers. The number of thread blocks that execute concurrently on an SM is referred to as the \textit{occupancy} of the kernel. Higher occupancy is typically desirable, as thread-level parallelism can be exploited to hide the latency of memory and arithmetic operations. GPUs have a large but high-latency global memory that is accessible to all SMs, an L2 cache that is shared by all SMs, and L1 caches that are local to each SM. When a warp of threads access global memory, GPUs try to coalesce the accesses into as few transactions as possible.

\section{Sparse Matrix Computation}

This section explains the operations implemented by our SpMM and SDDMM kernels.

\subsection{Sparse Matrix--Matrix Multiplication Operation}
Our SpMM kernel implements the computation $AB {\Rightarrow} C$, where $A$ is sparse and stored in the standard compressed sparse row (CSR) format. In the following sections, we refer to matrices $A$, $B$ and $C$ as the sparse matrix, dense matrix, and output matrix respectively. 

\subsection{Sampled Dense--Dense Matrix Multiplication Operation}
The SDDMM operation is defined as $AB \odot C {\Rightarrow} D$, where $C$ and $D$ are sparse and $\odot$ denotes the element-wise product of two matrices \cite{sddmm-hpml, adaptive-sparse-tiling}. Thanks to the element-wise scaling with a sparse matrix, dot-products for zero-valued locations of the output can be skipped to accelerate computation. 

In sparse deep neural networks, SDDMM is necessary for a number of key computations. For example, in a weight sparse neural network the forward pass computes $WX {\Rightarrow} Y$, where $W$ is sparse. In the backward pass, the gradient w.r.t. the sparse weights is computed as $\delta YX^T \odot \mathbb{I}[W]{\Rightarrow} \delta W$, where $\mathbb{I}[W]$ is an indicator function that returns 1 in the location of the nonzero values of the sparse matrix $W$. Transformer models with sparse attention similarly compute $QK^T \odot \mathbb{I}[Y]{\Rightarrow}Z$ in the forward pass, where $Q$ and $K$ are the query and key inputs to the attention mechanism respectively and $Y$ is a sparse matrix that describes the connectivity of the attention mechanism.

These computations differ from the strict definition of SDDMM in two ways. First, they do not require the element-wise scaling by the sparse matrix values. Secondly, the $B$ input operand to the SDDMM is typically transposed. With these applications in mind, our SDDMM implements the computation $AB^T \odot \mathbb{I}[C]{\Rightarrow}D$. While we specialize to the computation that arises in deep learning, we note that our approach is easily extensible to the general SDDMM computation \footnote{Element-wise scaling adds 1 load and 1 multiply instruction prior to storing the output. Non-transposed right-hand operand makes memory accesses trivially coalesced and simplifies the kernel relative to the transposed case.}.

\subsection{Data Organization}

To enable coalesced memory accesses into all input and output matrices, we store dense matrices in row-major layout and sparse matrices in CSR format \cite{cusparse, magma-spmm, spmm-design-principles}. We note that computing SpMM as $BA{\Rightarrow}C$, where $A$ is the sparse matrix stored in compressed sparse column format and $B$ and $C$ are stored column-major would be equally efficient.

\section{Sparse Matrix--Matrix Multiplication}

This section details the design of our SpMM kernel.

\subsection{Hierarchical 1-Dimensional Tiling}

Our scheme for SpMM on GPU is diagrammed in Figure \ref{tiling} and presented in CUDA pseudo-code in Figure \ref{pseudo_1d_tiling_spmm}. The decomposition follows a row-splitting scheme \cite{spmm-design-principles}, with one key difference: rather than mapping a thread block to an entire row of the output matrix, we shard the output into 1-dimensional tiles and map independent thread blocks to each. 

The motivation for this approach stems from the fact that the number of columns in the dense matrix can vary drastically in deep learning applications\footnote{Existing work on SpMM often focuses on problems where the dense matrix is "tall and skinny" \cite{cusparse, spmm-design-principles, magma-spmm}.}. Consider various neural network architectures with sparse weight matrices and dense activations. When training RNNs this dimension corresponds to the batch size, which is typically between 16-128 elements \cite{deepbench}. In Transformer architectures, this dimension is the product of the batch size and sequence length, which can vary from 256 to over 2048 elements \cite{transformer, image-transformer}. In 1$\times$1 convolutions, this dimension is the product of the image height and width with the batch size. In EfficientNet architectures, the product of the spatial dimensions alone range from 64 to 14,400 \cite{efficientnet}.

There are three main benefits to 1-dimensional tiling. Firstly, we can easily templatize our implementation for different tile sizes and generate specialized kernel variants for different regions of the problem space. Secondly, for problems with small M and K dimensions we launch more thread blocks than would otherwise be possible, enabling us to achieve higher occupancy and a higher fraction of peak throughput. Lastly, processing fixed-sized blocks enables us to aggressively unroll loops and compute offsets and constants at compile time. We similarly iterate through the reduction dimension in fixed-size steps, enabling further loop unrolling and static evaluation.

\subsection{Vector Memory Operations}

Vector memory instructions are an important tool for mitigating bandwidth bottlenecks and decreasing the number of instructions needed to express a computation \cite{vector-mem-ops}. However, it is non-trivial to use these operations in sparse matrix kernels. 

First, using vector memory instructions increases the number of values loaded simultaneously by a thread block. For example, a thread block with a single warp using 4-wide vector loads would request 128 floats with a single instruction. In our 1D tiling scheme, this means that some loads from a sparse matrix row of length less than 128 would need to be predicated off. Similarly, problems with fewer than 128 columns in the dense matrix would execute with some threads in every thread block predicated off for the entirety of the kernel's execution. These constraints limit the utility of vector memory accesses, applied naively, to very large problems.

Secondly, vector memory accesses require that the target values be aligned to the vector width (2 or 4 32-byte values). For accesses into the dense matrix or output matrix this requires that the number of columns be divisible by the vector width such that the start of every row of values is properly aligned. The larger issue is with loads from the sparse matrix. With a 1-dimensional tiling or row-splitting scheme, accesses within a thread block begin at the start of a row of values in the sparse matrix. Because rows in a sparse matrix can have arbitrary lengths, these initial addresses have no alignment guarantees regardless of the problem dimensions.

\usemintedstyle{borland}
\renewcommand{\theFancyVerbLine}{\textcolor[RGB]{0,0,0}{\small \arabic{FancyVerbLine}}}
\begin{figure}[t]
\begin{minted}[
fontfamily=courier,
fontsize=\footnotesize, 
xleftmargin=14pt, 
numbersep=4pt, 
linenos, 
frame=lines]{cuda}
template <int kBlockItemsK, int kBlockItemsX>
__global__ void SpmmKernel(
  SparseMatrix a, Matrix b, Matrix c) {
  // Calculate tile indices.
  int m_idx = blockIdx.y;
  int n_idx = blockIdx.x * kBlockItemsX;
  
  // Calculate the row offset and the number 
  // of nonzeros in this thread block's row.
  int m_off = a.row_offsets[m_idx];
  int nnz = a.row_offsets[m_idx+1] - m_off;  
    
  // Main loop.
  Tile1D c_tile(/*init_to=*/0);
  for(; nnz > 0; nnz -= kBlockItemsK) {
    Tile1D a_tile = LoadTile(a);
    Tile2D b_tile = LoadTile(b);
    c_tile += a_tile * b_tile;
  }
  
  // Write output.
  StoreTile(c_tile, c);
}
\end{minted}
 \caption{\textbf{CUDA pseudo-code for SpMM with 1-dimensional tiling.} The output matrix is statically partitioned into 1-dimensional tiles. Independent thread blocks are launched to compute each output tile. On each iteration of the main loop, we load a 1-dimensional strip of the sparse matrix and a 2-dimensional tile of the dense matrix and accumulate the vector-matrix product. After processing all nonzero values in the row, the results are written to the output matrix.}
\label{pseudo_1d_tiling_spmm}
\end{figure}

\subsubsection{Subwarp Tiling}

To address the first issue, we extend our scheme to allow mapping of subsets of a warp (i.e., a subwarp) to independent 1D tiles of the output. This reduces the access width constraint by a factor of the number of subwarps used. This also gives us the flexibility to spread threads across more rows of the output matrix for problems with a smaller number of columns in the dense and output matrices.

With subwarp tiling, our scheme bears some resemblance to a standard two-dimensional tiling scheme at the warp level. The important difference is that subwarps processing different rows of the output matrix are not able to reuse values loaded from the dense matrix. However, depending on the sparsity level, accesses issued by different subwarps are likely to exhibit locality that could be serviced through caches.

The main drawback to this approach is that rows of variable length can result in warp divergence. We address the issue of load imbalance between threads in a warp in section \ref{load_balancing}.

\subsubsection{Reverse Offset Memory Alignment}

A simple approach to address the second issue is to pad the rows of the sparse matrix with zeros such that all rows are a multiple of four in length. However, this limits the generality of the kernel. To enable the use of vector memory instructions on arbitrary sparse matrices, we introduce a simple trick in the setup portion of the kernel (AKA, the prelude): after loading the row offset and calculating the row length, each thread block decrements its row offset to the nearest vector-width-aligned address and updates the number of nonzeros that it needs to process. To maintain correctness, the threads mask any values that were loaded from the previous row prior to accumulating the result in the first iteration of the main loop.

We refer to this trick as \textit{reverse offset memory alignment} (ROMA). Relative to the explicit padding scheme, ROMA does not change the amount of work done by each thread block. The key difference is that instead of explicitly padding the matrix data structure, ROMA effectively pads the rows of the sparse matrix with values from the row before it\footnote{Note that the first row of the sparse matrix is guaranteed to be vector aligned, as all CUDA memory allocation routines allocate memory with at least 256-byte alignment \cite{cuda-programming-guide}}. 

ROMA can be implemented very efficiently. The alignment process adds 6 PTX instructions in the kernel prelude: 2 bitwise \verb!and!, 1 \verb!add!, 1 \verb!setp! (set predicate), and 2 \verb!selp! (select based on predicate). The masking process adds 1 \verb!setp! and 2 \verb!str.shared! (shared memory store) instructions to the first iteration of the main loop.

\begin{figure}[t]
  \includegraphics[width=\columnwidth]{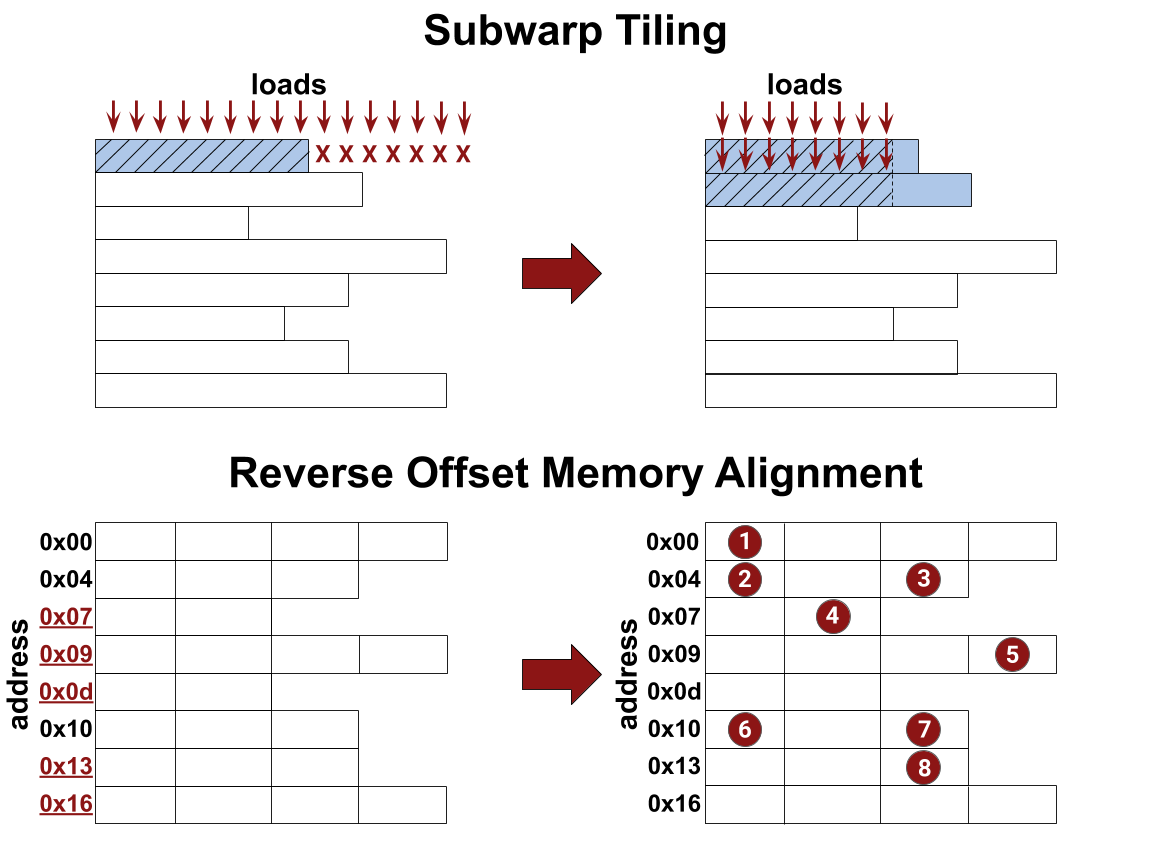}
  \vspace{-6mm}
  \caption{\textbf{Techniques for enabling the use of vector memory operations on sparse matrix data structures}. \textbf{Top:} Subwarp tiling maps subsets of a warp to independent 1-dimensional tiles of the output. By splitting warp accesses across multiple rows we reduce the amount of wasted work. Rows assigned to a warp/subwarp are marked in blue. Load addresses are indicated with arrows and predicated loads are indicated with "X". The hashed region denotes the values loaded by a single set of loads across the warp/subwarp. \textbf{Bottom:} Reverse offset memory alignment backs up the address of each row to the nearest aligned address. Values from the previous row are masked in the first loop iteration to maintain correctness. Misaligned addresses are underlined. Modified row start addresses are marked with circles and row index.}
\label{vector_memory_ops}
\end{figure}

These techniques for enabling the use of vector memory instructions on sparse matrices are visualized in Figure \ref{vector_memory_ops}. Figure \ref{pseudo_subwarp_spmm} shows CUDA pseudo-code for our SpMM kernel with the necessary modifications for subwarp tiling and ROMA.

\subsection{Row Swizzle Load Balancing}
\label{load_balancing}

A number of approaches for handling load imbalance in sparse computation have been proposed \cite{merge-based-spmv, spmm-design-principles, adaptive-sparse-tiling}. However, existing approaches tightly couple load balancing to the parallelization scheme. While these schemes achieve good performance when load imbalance is significant, they typically introduce computational irregularity that can damage performance on more regular problems problems \cite{spmm-design-principles}. However, despite the regularity of sparse matrices found in DNNs, our kernels still suffer from load imbalance (see Figure \ref{load_balancing_ablation}).


When mapping sparse matrix operations to GPUs, there are two potential sources of load imbalance \cite{spmm-design-principles}.

\begin{enumerate}[label=(\alph*)]
\item \textbf{Load imbalance between warps or thread blocks:} Some warps/thread blocks may be assigned less work than others. This can lead to some SMs sitting idle while others do useful work.
\item \textbf{Load imbalance within a warp or thread block:} Some threads within a warp may be assigned less work than others. This can lead to warp divergence and inefficient use of math units and memory bandwidth within an SM.
\end{enumerate}

To address these issues, we make two observations. First, many units of work of varying sizes will be scheduled onto each SM over the course of kernel execution. Secondly, we can control what work is assigned to which SM by changing which threads are assigned to process each unit of work. Based on these observations, we can ensure the workload is balanced across the processing elements by remapping where work is scheduled such that each processing element is assigned a roughly equal amount of work. We refer to this remapping procedure as a \textit{row swizzle}\footnote{In reference to the similar approach of thread block swizzles, where thread blocks are reordered to improve cache reuse \cite{cutlass} as well as the fact that the reordering in our case is at the granularity of an output row.}. To address both sources of load imbalance, we introduce a two level re-ordering of work:

\begin{enumerate}[label=(\alph*)]
\item \textbf{Row Binning:} Given an understanding of how thread blocks are mapped to SMs, alter the tile mappings such that each SM receives approximately the same amount of work to do. This helps to address load imbalance between warps/thread blocks.
\item \textbf{Row Bundling:} For kernels where warps are split across multiple rows of the sparse matrix, alter the tile mappings such that each subwarp receives approximately the same amount of work to do. This helps address load imbalance within a warp.
\end{enumerate}

\subsubsection{Volta Thread Block Scheduler}
The binning of rows such that SMs receive roughly the same amount of work is complex to implement, as it depends on the GPUs thread block scheduling algorithm, which is not public knowledge. We reverse engineer the Nvidia Volta thread block scheduler, following the same general approach as \cite{fermi-tbs}.

Overall, the Volta thread block scheduler is much simpler than the Fermi thread block scheduler. Thread blocks in the first wave are assigned to SMs based on their block index:

\vspace{-4mm}
$$sm\_idx = 2(block\_idx \bmod 40) + \frac{block\_idx}{40} \bmod 2$$

where $block\_idx$ is calculated:

\vspace{-4mm}
$$block\_idx = blockIdx.x + blockIdx.y * gridDim.x$$

This mapping distributes thread blocks round-robin over the SMs. After the first wave, thread blocks are scheduled in order of $block\_idx$ as resources become available.

\subsubsection{Row Binning \& Row Bundling Heuristics}

A simple heuristic for binning rows is to select the first wave to be the heaviest N row bundles and then pair the following N heaviest row bundles with the previous bundles in reverse order of heaviness. To bundle the rows by size, we can greedily create bundles from consecutive rows ordered by size.

Given the online thread block scheduling algorithm used by Nvidia GPUs, these two heuristics can be implemented with a sort of the row indices by row length. Given a sorted array of the row indices in order of decreasing size, bundles consist of blocks of consecutive row indices. The first wave of bundles are scheduled round-robin across the SMs, and remaining bundles are scheduled in decreasing order of heaviness as bundles complete execution. We note that this heuristic for row binning is similar to guided self-scheduling \cite{guided-self-scheduling}.

An advantage of this approach is that we do not need to know the target bundle size to group similarly sized rows. This means that the heuristic does not need to have insight into any kernel selection heuristics used under the hood. 

Since the topology of sparse matrices in DNNs is typically updated infrequently, the cost of the \verb!argsort! of the row indices by their row lengths can be amortized over many training steps \cite{blocksparse-gpu-kernels, tsos, sparse-transformer}. Implementing the swizzle in the kernel also requires the addition of a single load during the kernel prelude. The memory required to store the sorted indices for the matrix is negligible, as the number of rows in the matrix is typically much smaller than the number of nonzeros in the matrix for our target applications.

\begin{figure}[t]
  \includegraphics[width=\columnwidth]{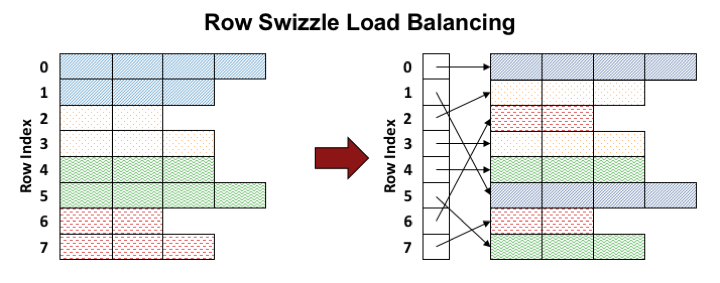}
  \vspace{-8mm}
  \caption{\textbf{Row swizzle approach for load balancing sparse matrix computation}. Rows processed by the same warp are marked with the same pattern and color. We introduce a layer of indirection that re-orders when rows are processed. To balance work between threads in a warp, we group rows of similar length into \textit{bundles}. To balance work between SMs, we process row bundles in decreasing order of size.}
\label{load_balancing_diagram}
\end{figure}

\begin{figure}[t]
  \includegraphics[width=\columnwidth]{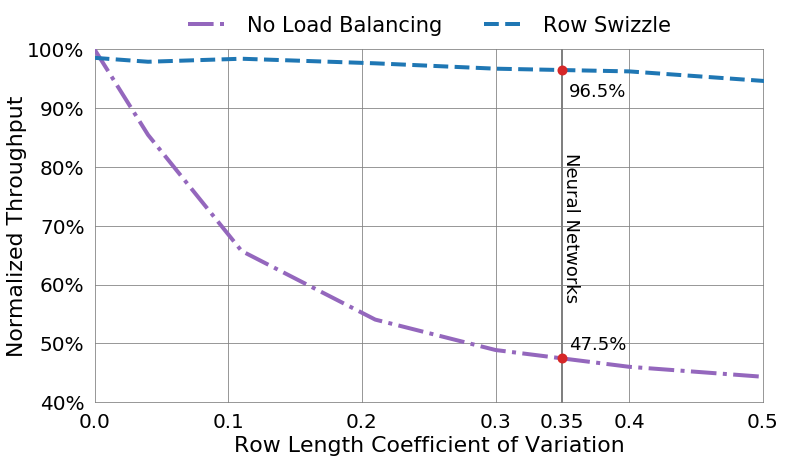}
  \vspace{-6mm}
  \caption{\textbf{Sparse matrix--matrix multiplication runtime with varying levels of load imbalance}. M=8192, K=2048, N=128, sparsity=75\% in single-precision on an Nvidia V100 GPU. Throughput measured as a percent of the throughput achieved with a sparse matrix where all rows have the same number of nonzero values. The gray line labeled "Neural Networks" marks the average CoV of sparse matrices in our dataset of deep neural networks. With this problem configuration, performance of the standard row ordering degrades to 47.5\% of throughput with a perfectly balanced sparse matrix. Our row swizzle load balancing technique maintains 96.5\% of the throughput with a perfectly balanced sparse matrix. We document the performance of our row swizzle load balancing technique further in Section \ref{ablation_study}}
\label{load_balancing_ablation}
\end{figure}

Figure \ref{load_balancing_diagram} shows the high-level scheme for row swizzle load balancing. Figure \ref{load_balancing_ablation} shows the performance of row swizzle load balancing for a sample problem as load imbalance increases. Figure \ref{pseudo_subwarp_spmm} shows CUDA pseudo-code for our SpMM kernel with the necessary modifications for row swizzle load balancing. 

\subsection{Implementation Details}

This section details additional low-level optimizations we applied to achieve good performance.

\subsubsection{Index Pre-Scaling}
In each iteration of the main loop of our kernel, we load the sparse matrix values and indices and store them in shared memory. Each index will be used by all threads to load from the dense matrix. To avoid redundant work each time an index is loaded, we have each thread scale its portion of the indices prior to storing to shared memory. 


\subsubsection{Residue Handling}

Our kernel processes as many full tiles of nonzero values as possible and then executes a residue handler to accumulate the remaining products. As sparse matrix row lengths are rarely divisible by the tile size, it's important that the residue handling code is highly efficient.

To maximize shared memory bandwidth and minimize bank conflicts, we use 128-bit shared memory load instructions whenever possible \cite{sparse-persistent-rnns}. This is trivial for the main loop, but difficult for the residue handling code as the number of nonzeros remaining is not necessarily divisible by four. To enable the use of wide shared memory instructions, we zero the shared memory buffers used for sparse matrix values and indices prior to loading the residual values and indices. We then split the loops for dense matrix loading and computation into two, and unroll the inner loop 4$\times$ without bounds checks. 


\subsubsection{Mixed Precision}

In addition to standard 32-bit floating-point kernels, we extended our SpMM implementation to support mixed-precision computation, as is commonly used in deep learning \cite{mixed-precision-training}. Our kernels support 16-bit floating-point input/output data and use 16-bit integer indices for the sparse matrix meta-data. Inside our kernel, we convert FP16 data to FP32 and issue FP32 fused multiply-add instructions, as is standard. We convert the final outputs from FP32 to FP16 before writing the result. Due to the reduced representational capacity of 16-bit integers, we do not perform our index pre-scaling optimization for mixed-precision kernels.

\section{Sampled Dense--Dense Matrix Multiplication}

This section details the design of our SDDMM kernel.

\subsection{Hierarchical 1-Dimensional Tiling}

We use the same 1D tiling scheme for SDDMM as we do for SpMM, with two main differences. First, instead of mapping thread blocks to 1D regions of the output we map them to 1D strips of consecutive nonzeros. Because the output is sparse, this ensures better work distribution across thread blocks and is simpler to implement. Because the number of nonzeros in each row cannot be inferred without inspecting the sparse matrix, we launch the maximum number of thread blocks that could be needed. On startup, each thread block calculates if it has work to do and returns early if it is not needed. An alternative approach would be to use dynamic parallelism \cite{cuda-dynamic-parallelism}. However, we do not observe significant overhead from launching extra thread blocks in our benchmarks. For SDDMM targeting problems with very high sparsity, it's possible that dynamic parallelism would lead to better performance.

The second difference in our work decomposition is caused by the need to perform the computation with the transpose of the right-hand operand. With the dense matrices stored in row-major layout, naively partitioning the outputs across the threads would result in strided, uncoalesced memory accesses to the right-hand matrix. To avoid this issue, we alter our scheme so that each thread mapped to an output tile computes a portion of the results for all outputs in that tile. We then perform a reduction between these threads using warp shuffle instructions to compute the final results for each thread. 

\definecolor{hcolor}{HTML}{AEC7E8}
\begin{figure}[t]
\begin{minted}[
fontsize=\footnotesize, 
xleftmargin=14pt, 
numbersep=4pt, 
linenos, 
frame=lines, 
highlightlines={5-8, 11-13, 20-24, 30}, 
highlightcolor=hcolor
]{cuda}
template <int kBlockItemsY, int kBlockItemsK, 
  int kBlockItemsX>
__global__ void SpmmKernel(
  SparseMatrix a, Matrix b, Matrix c) {
  // Subwarp tiling: calculate tile indices 
  // for this subwarp.
  int m_idx = blockIdx.y * kBlockItemsY + 
              threadIdx.y;
  int n_idx = blockIdx.x * kBlockItemsX;
    
  // Row swizzle: load this subwarp's row 
  // index from the pre-ordered indices.
  m_idx = a.row_indices[m_idx];
  
  // Calculate the row offset and the number 
  // of nonzeros in this thread block's row.
  int m_off = a.row_offsets[m_idx];
  int nnz = a.row_offsets[m_idx+1] - m_off;  
  
  // ROMA: align the row pointer so that we 
  // can use vector memory instructions.
  MemoryAligner aligner(m_off, nnz);
  nnz = aligner.AlignedNonzeros();
  m_off = aligner.AlignedRowOffset();

  // First loop iteration.
  Tile1D c_tile(/*init_to=*/0);
  if (nnz > 0) {
    Tile1D a_tile = LoadTile(a);
    aligner.MaskPrefix(a_tile);
    Tile2D b_tile = LoadTile(b);
    c_tile += a_tile * b_tile;
    nnz -= kBlockItemsK;
  }
  
  // Main loop.
  for(; nnz > 0; nnz -= kBlockItemsK) {
    Tile1D a_tile = LoadTile(a);
    Tile2D b_tile = LoadTile(b);
    c_tile += a_tile * b_tile;
  }
  
  // Write output.
  StoreTile(c_tile, c);
}
\end{minted}
\vspace{-3mm}
\caption{\textbf{CUDA pseudo-code for SpMM with subwarp tiling, ROMA and row swizzle load balancing.} To enable the use of vector memory instructions on a wider range of problem configurations, subsets of a warp are mapped to 1-dimensional output tiles, adding a template parameter that denotes the number of subwarps used (lines 1-2) and altering the index calculations (lines 5-8). To decrease load imbalance, we alter the order in which rows are processed (lines 11-13). To enable the use of vector memory instructions on misaligned addresses in sparse matrices, we back-up the row pointer to the nearest aligned address (lines 20-24). To maintain correctness, we mask any values loaded from the previous row in the first iteration of the main loop (line 30).}
\label{pseudo_subwarp_spmm}
\end{figure}

\begin{figure*}[t]
  \includegraphics[width=\textwidth]{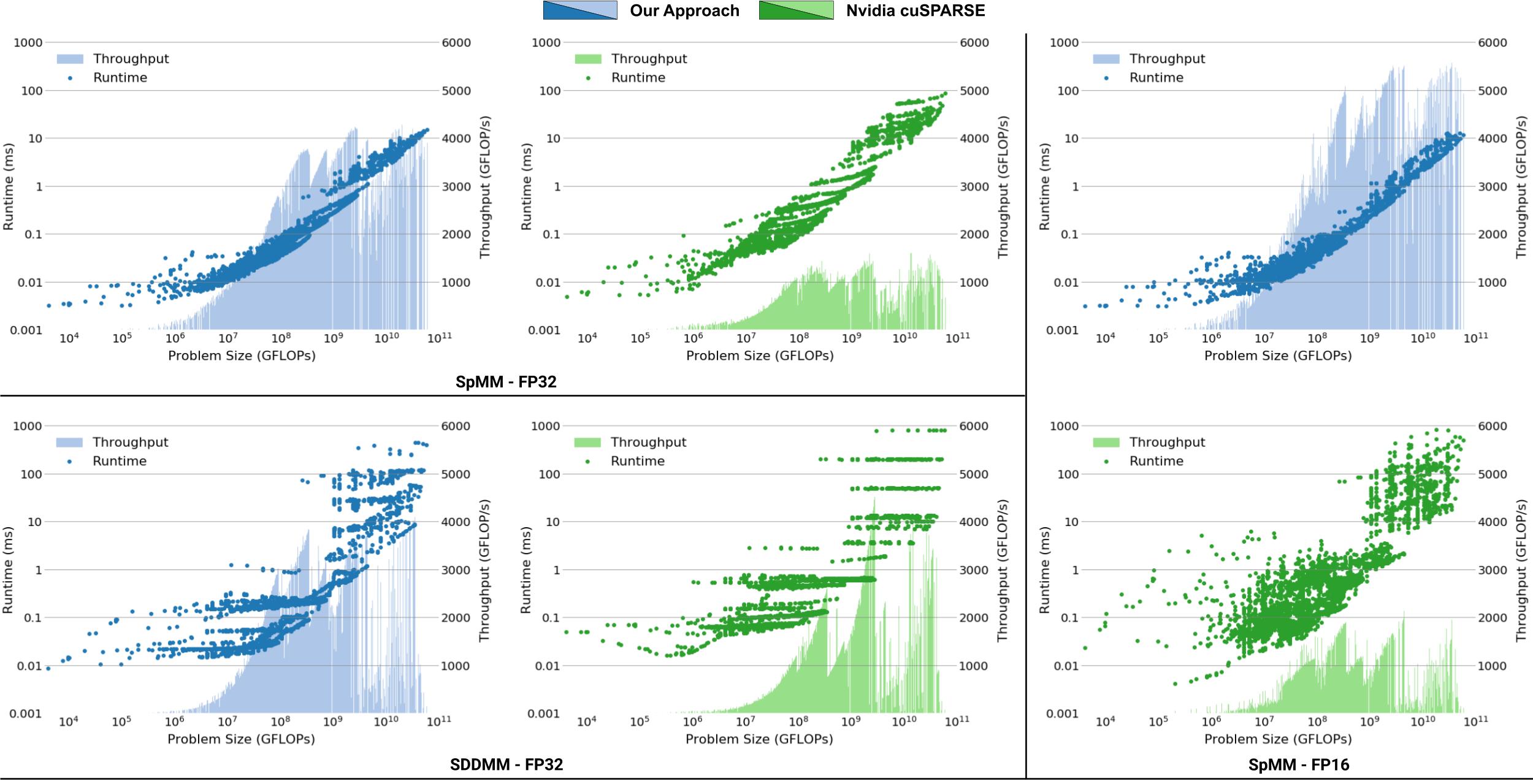}
  \vspace{-6mm}
  \caption{\textbf{Benchmarks on our dataset of sparse matrices from deep neural networks.} Runtime (left y-axis) and throughput (right y-axis) plotted with increasing problem size for each kernel and precision. Benchmarked on an Nvidia V100 GPU. \textbf{Top Left:} SpMM benchmarks in single-precision. Across all problems, our approach achieves a geometric mean speedup of 3.58$\times$ and a peak speedup of 14.2$\times$ over Nvidia cuSPARSE. \textbf{Bottom Left:} SDDMM benchmarks in single-precision. Across all problems, our approach achieves a geometric mean speedup of \rev{2.19$\times$} and a peak speedup of \rev{6.58$\times$} over Nvidia cuSPARSE. \textbf{Right:} SpMM benchmarks in mixed precision with 16-bit data and 32-bit computation. Across all problems, our approach achieves a geometric mean speedup of 5.97$\times$ and a peak speedup of 297.5$\times$ over Nvidia cuSPARSE.
}
\label{dnn_benchmarks}
\end{figure*}

\begin{figure*}[t]
  \includegraphics[width=\textwidth]{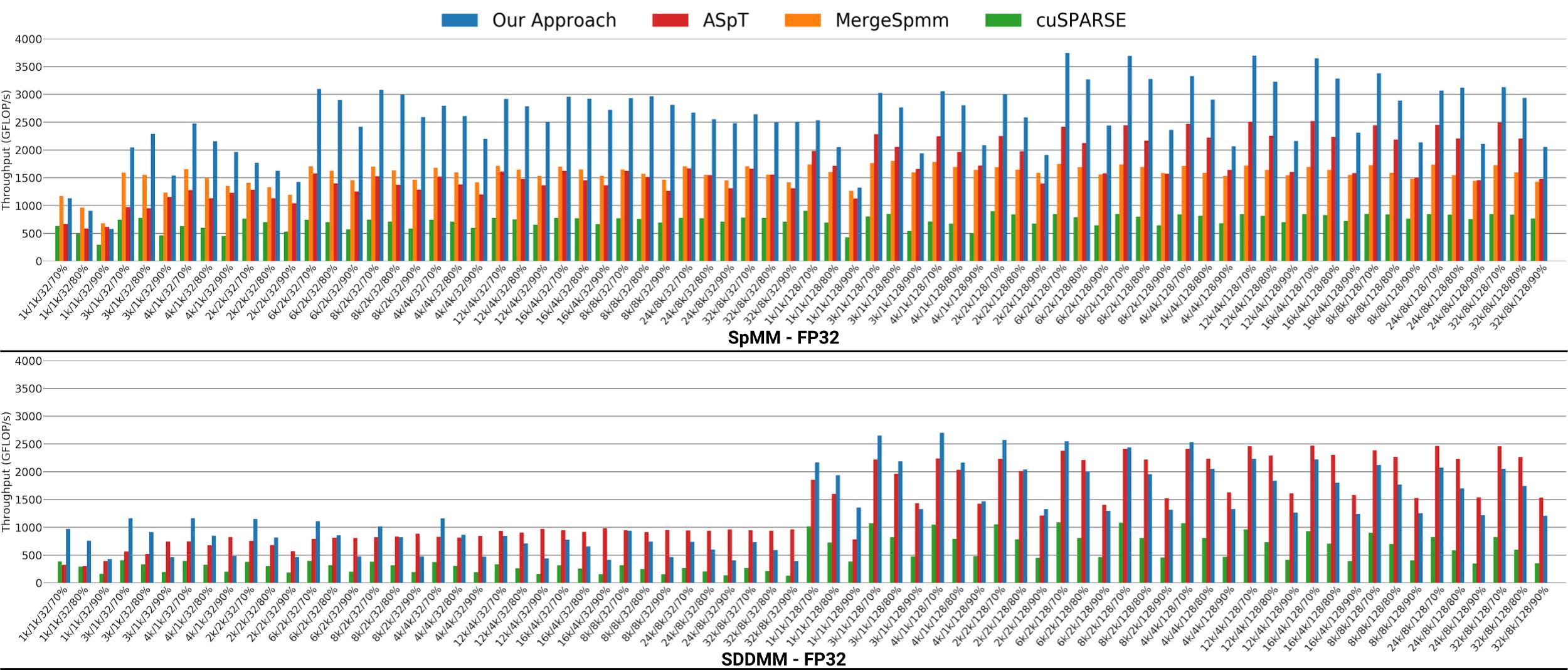}
  \vspace{-6mm}
  \caption{\rev{\textbf{Benchmarks on sparse recurrent neural neural network problems.} 
  Each problem is labeled M/K/N/sparsity. All benchmarks taken on an Nvidia V100 GPU in single-precision. \textbf{Top:} SpMM benchmarks. Compared to ASpt \cite{adaptive-sparse-tiling}, our kernel achieves a geometric mean speedup of 1.56$\times$ and a peak speedup of 2.4$\times$. Compared to the merged-based approach of \cite{spmm-design-principles}, our kernel achieves a geometric mean speedup of 1.59$\times$ and a peak speedup of 2.15$\times$. Compared to cuSPARSE, our kernel achieves a geometric mean speedup of 3.47$\times$ and a peak speedup of 4.45$\times$. \textbf{Bottom:} SDDMM benchmarks. Our kernel performs competitively with the adaptive sparse tiling approach of \cite{adaptive-sparse-tiling}, achieving 92\% of the throughput on average while using 3$\times$ less memory and no-reordering of the sparse matrix. Compared to  cuSPARSE, our kernel achieves a geometric mean speedup of 2.69$\times$ and a peak speedup of 3.51$\times$.}}
\label{rnn_benchmarks}
\end{figure*}

An alternative to this approach would be to perform the transpose of values loaded from the right-hand matrix in shared memory prior to computation. While this would use less registers per-thread, it would double shared memory usage. On Nvidia Volta GPUs shared memory and L1 cache use the same storage. Thus, using more shared memory reduces the size of the L1 cache. For these kernels, we found L1 cache capacity to be important for performance and thus decided against performing an explicit shared memory transpose.

\subsection{Vector Memory Operations}
Because both inputs are dense, it is trivial to use vector loads/stores for SDDMM problems where the inner dimension is divisible by the vector width. For all problems, we use scalar loads/stores on the sparse matrix. These operations only occur at the beginning and end of the kernel and do not significantly affect performance. To enable the use of vector loads/stores on a wider range of problems we process output tiles with subwarps, as explained in the context of SpMM.

\subsection{Implementation Details}

While we do use subwarp tiling to enable the use of vector memory instructions on a wider range of problems, load balancing in SDDMM is less critical due to the fact that all dot-products to be computed are of equal length. Additionally, problems from deep neural networks commonly have a dot-product length that is divisible by the SIMT width, making efficient residue handling less critical than in SpMM. For the SDDMM residual computation we use the same loop structure as the main loop and do not apply our loop-splitting optimization to enable wide shared memory loads.

\section{Experiments}
\label{experiments}
This section provides empirical results and analysis of our SpMM and SDDMM kernels. For SpMM we use a kernel selection heuristic where we select the n-dimension tile size to be $N$, rounded up to a power of 2, up to a maximum of 64. For SDDMM we use an n-dimension tile size of 32. For both kernels we use the widest vector memory operations possible. \rev{All benchmarks were conducted with CUDA 10.1.}

\subsection{Kernel Benchmarks}
\label{kernel_benchmarks}

\subsubsection{Sparse Matrix Dataset}
We evaluate the performance of our SpMM and SDDMM kernels by benchmarking on our dataset of sparse matrices from deep neural networks. For each of the 3,012 matrices in the dataset, we benchmark with both training and inference batch sizes. For SDDMM, we benchmark the problem corresponding to the gradient with respect to the sparse weight matrix. We benchmark convolution operations found in ResNet-50, as an \verb!im2col! transform on the input data followed by SpMM or SDDMM\cite{cudnn}. We do not include the time of the \verb!im2col! transform in our benchmarks. For ResNet-50 benchmarks with inference batch size, we pad the batch dimension to the nearest multiple of four to enable vector memory instructions. All benchmarks are performed on an Nvidia V100 GPU. We use Nvidia cuSPARSE's \verb!cusparseSpMM! and \verb!cusparseConstrainedGeMM! as the baselines for our SpMM and SDDMM benchmarks respectively. Both cuSPARSE kernels use column-major layouts for dense matrices and CSR format for sparse matrices. Because \verb!cusparseConstrainedGeMM! does not support transposition of the right-hand operand, we explicitly transpose the matrix using cuBLAS and include the transposition in our timing. We benchmark all problems on a single Nvidia V100 GPU. The results of all benchmarks are presented in Figure \ref{dnn_benchmarks}. We present statistics for these benchmarks in Table \ref{dnn_benchmarks_table}.

Across all benchmarks, our SpMM and SDDMM kernels show significant advantages over Nvidia cuSPARSE. For single-precision SpMM, our kernel achieves a geometric mean speedup of 3.58$\times$ and reaches 4.29TFLOPs, representing 27.3\% of single-precision peak. Our kernel outperforms cuSPARSE on 99.75\% of the problems in our dataset. For single-precision SDDMM, our kernel achieves a geometric mean speedup of \rev{2.19$\times$} and reaches \rev{4.11TFLOPs}, representing \rev{26.2\%} of single-precision peak. Our kernel outperforms cuSPARSE on \rev{93.34\%} of the problems in our dataset.

With mixed-precision, our SpMM kernel achieves a geometric mean speedup of 5.97$\times$ over Nvidia cuSPARSE and a peak throughput of 5.57TFLOPs. While our kernel uses 16-bit integers for the sparse matrix meta-data, cuSPARSE only supports 32-bit indices. We find it likely that this contributes to the increased performance gap. Our mixed-precision SpMM outperforms cuSPARSE on 99.7\% of the problems in our dataset. We note that cuSPARSE's mixed-precision SpMM performs inconsistently on some problems, leading to extreme slowdowns of as much as 297.5$\times$ relative to our kernel.

\subsubsection{\rev{Sparse Recurrent Neural Networks}}

\rev{This section evaluates the performance of our kernels relative to the recently proposed techniques of \cite{spmm-design-principles} and \cite{adaptive-sparse-tiling}. The SpMM kernel provided by \cite{spmm-design-principles} only supports problems with batch sizes divisible by 32. \cite{adaptive-sparse-tiling} wrote SpMM and SDDMM kernels for batch size 32 and 128 and also require that the number of rows in the sparse matrix be divisible by 256. Given these constraints, we opt to benchmark these kernels on a dataset of problems from recurrent neural networks, where the problem configurations supported by the kernels from \cite{spmm-design-principles} and \cite{adaptive-sparse-tiling} are realistic for deep neural networks. We benchmark each kernel on RNN, gated recurrent unit (GRU) \cite{gru}, and long short-term memory network (LSTM) \cite{lstm} problems with sparse weights. We generated sparse matrices with random uniform sparsity. We benchmarked problems with state sizes 1k, 2k, 4k, and 8k, sparsities 70\%, 80\%, and 90\% and batch sizes 32 and 128. All benchmarks were performed on an Nvidia V100 in single-precision. We do not include the time require for the pre-processing step used by the Adaptive Sparse Tiling (ASpT) approach of \cite{adaptive-sparse-tiling} in our benchmarks. We benchmark the row-splitting kernel from \cite{spmm-design-principles}, as all of our benchmarks are beyond the threshold of average row length that the authors use to select between their row-splitting and nonzero-splitting kernels. Benchmark results are presented in Figure \ref{rnn_benchmarks}.}

\rev{For SpMM, our approach significantly outperforms other methods. Our approach achieves geometric mean speedups of 1.56$\times$, 1.59$\times$, and 3.47$\times$ over MergeSpmm\cite{spmm-design-principles}, ASpT, and cuSPARSE respectively. For SDDMM, our approach significantly outperforms cuSPARSE and achieves performance on-par with ASpT. Our approach achieves geometric mean speedups of 2.69$\times$ over cuSPARSE and 92\% of the throughput of ASpT on average. While ASpT achieves good performance for SDDMM, it has a number of limitations. First, including the original CSR matrix, ASpT requires 3$\times$ the memory to store the re-ordered matrix as well as meta-data needed for tiled execution. 
Second, the author's implementation uses different re-orderings of the sparse matrix for SpMM and SDDMM problems. For deep learning applications, this means that gradients calculated with respect to a sparse matrix will be in a different order than the sparse matrix used in the forward pass. In order to perform gradient updates or continue backpropagation, applications must pay the cost of re-ordering the sparse matrix on every training iteration.}

\subsection{Ablation Study}
\label{ablation_study}
Table \ref{ablation_study_table} shows the results of our ablation study on the optimizations we propose for each kernel. We benchmark both kernels on our dataset of sparse matrices from DNNs with both training and inference batch sizes. We report statistics for each model and batch size separately to show the effect of each technique on different portions of the problem space.

Across these benchmarks we find that techniques like row swizzle load balancing and residue unrolling are robust to varying problem configurations, while vector memory instructions show large benefits for compute heavy problems and less benefit for small problems. One outlier is the superior performance of scalar memory operations for SDDMM. \rev{With the small weight matrices found in these models, these problems are largely occupancy-bound and thus benefit from the fact that our scalar kernels process fewer outputs per thread. On the dataset of RNN problems studied in Section \ref{rnn_benchmarks}, we observe our vector kernel achieve a geometric mean speedup of 2.45$\times$ over the scalar variants. These results indicate that better kernel selection heuristics could greatly improve performance.}

In addition to these techniques, our kernels benefit from the use of favorable data layouts and an efficient implementation enabled by our 1D tiling scheme.

\subsection{Application: Sparse Transformer}

Transformer models are a popular sequence modeling architecture, having been used to achieve state-of-the-art results on tasks such as machine translation \cite{transformer}, language modeling \cite{transformer-xl}, and image generation \cite{image-transformer}. Transformer models are made up of stacked layers, each of which contains a multi-head attention mechanism followed by a small fully-connected network. The attention mechanism used in Transformer takes in a query $Q$, key $K$, and value $V$ and computes a weighted average of the input values based on the similarity of $Q$ and $K$:

\vspace{-3mm}
$$Attention(Q, K, V) = Softmax(\frac{QK^T}{\sqrt{d_k}})V$$
\vspace{-4mm}

Where $d_k$ is the number of features for each element of the sequence. Despite the effectiveness of this architecture, $QK^T$ computes the similarity of each token in the sequence with all other tokens, requiring computation and memory that grows quadratically with the sequence length. To alleviate this issue, recent work has explored the use of sparsity in the attention mechanism \cite{sparse-transformer, reformer, routing-transformer}. With sparse attention, we compute a subset of the outputs of $QK^T$ and then multiply the sparse output by $V$. With unstructured sparsity, these operations correspond to an SDDMM followed by an SpMM.

\begin{table}[t]
\caption{\textbf{Sparse matrix dataset benchmark results.} \hspace{\columnwidth} Speedups reported relative to Nvidia cuSPARSE.}
\label{dnn_benchmarks_table}
\centering

\begin{tabular}{c|cc|c}
\multirow{2}{*}{\textbf{Kernel}} & \multicolumn{2}{c|}{\textbf{Single-Precision}} & \textbf{Mixed-Precision} \\
& SpMM & SDDMM & SpMM \\ \hline
Geo. Mean Speedup & 3.58$\times$ & \rev{2.19$\times$} & 5.97$\times$ \\
Peak speedup & 14.2$\times$ & \rev{6.58$\times$} & 297.5$\times$ \\
Peak throughput (TFLOPs) & 4.29 & \rev{4.11} & 5.57 
\end{tabular}
\end{table}

\subsubsection{Experimental Setup}
We trained a Transformer with sparse attention on the ImageNet-64x64 image generation dataset which has a sequence length of 12,288. Our model consists of 3 layers with 8 attention heads each, a hidden dimension of 1,024 and a filter size of 4,096 in the fully-connected network. We trained our models with a batch size of 8 for 140,000 training steps. For our sparse model, we simulate sparsity during training and convert to a sparse representation for benchmarking. While we train on an image generation task, we note that this architecture can be applied to other sequence learning tasks like language modeling without modification.

\begin{figure}[t]
  \includegraphics[width=\columnwidth]{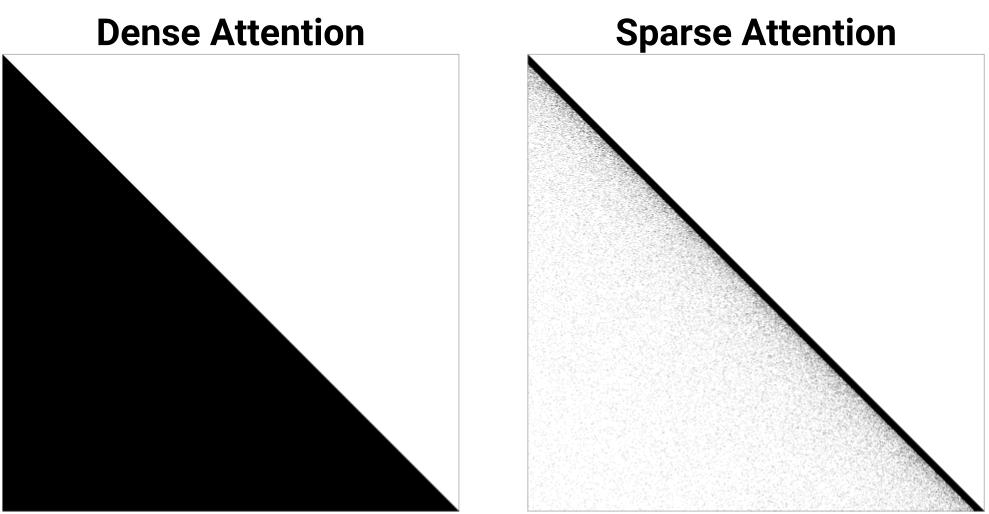}
  \vspace{-6mm}
  \caption{\textbf{Transformer attention mechanism connectivity.} The upper diagonal is masked so that tokens only attend to those that came before them. \textbf{Left:} Dense all-to-all attention.  \textbf{Right:} Sparse attention with a small dense band and random off-diagonal sparsity sampled with probability inversely proportional to distance from the diagonal.
  }
\label{sparse_attention}
\end{figure}

For our sparse model, we generated attention masks with a dense band of size 256 along the diagonal and random sparsity off-diagonal sampled with probability inversely proportional to the distance from the diagonal. We set off-diagonal sparsity to 95\%. The sparse attention mask stays the same over the course of training and is shared by all attention heads and layers. The attention mask used by our model is visualized in Figure \ref{sparse_attention}. We additionally wrote a kernel that computes the softmax function on a sparse matrix. For each model, we benchmark the forward pass in single-precision. 

\subsubsection{Results \& Analysis}

Benchmark results are reported in Table \ref{sparse_transformer}. On a V100 GPU, our sparse model achieves a 2.09$\times$ speedup and 12.8$\times$ memory savings over the standard Transformer while matching accuracy. We report accuracy in bits per dimension, as is standard for this task. Note that lower bits per dimension is desirable. In addition to our results on V100, we exploit the memory savings of our sparse model to benchmark on an Nvidia 1080. On a significantly less powerful GPU, our sparse model is able to process 32,039 tokens per second while the standard Transformer runs out of memory. The memory savings of the sparse Transformer could also be used to train a much larger model, leading to higher accuracy.

\setlength{\tabcolsep}{4pt}
\begin{table}[t]
\caption{\textbf{Ablation study for our SpMM and SDDMM kernels.} Performance measured as a percent of the performance of our complete kernels, averaged across all benchmarks.}
\label{ablation_study_table}
\centering

\begin{tabular}{c|cccc}
\hline
\multicolumn{5}{c}{\textbf{SpMM}} \\ \hline
\textbf{Model} & Transformer & Transformer & ResNet-50 & ResNet-50 \\ 
\textbf{Batch Size} & 1 & 8 & 1 & 256 \\ \hline
\textbf{-Load Balancing} & 96.1\% & 88.9\% & 91.7\% & 78.5\% \\
\textbf{-Vector Inst.} & 100.1\% & 80.9\% & 87.9\% & 64.8\% \\
\textbf{-Residue Unroll} & 92.0\% & 94.1\% & 87.8\% & 92.6\% \\
\textbf{-Index Pre-Scale} & 100.6\% & 100.6\% & 98.2\% & 100.3\%               
\end{tabular}
\vspace{2mm}

\begin{tabular}{c|cccc}
\hline
\multicolumn{5}{c}{\textbf{SDDMM}} \\ \hline
\textbf{Model} & Transformer & Transformer & ResNet-50 & ResNet-50 \\ 
\textbf{Batch Size} & 1 & 8 & 1 & 256 \\ \hline
\textbf{-Load Balancing} & \rev{101.1\%} & \rev{97.1\%} & \rev{100.9\%} & \rev{96.8\%} \\
\textbf{-Vector Inst.} & \rev{98.3\%} & \rev{132.0\%} & \rev{120.2\%} & \rev{170.6\%}
\end{tabular}
\end{table}

\subsection{Application: Sparse MobileNetV1}
MobileNetV1 is an efficient convolutional neural network for computer vision tasks \cite{mobiletnetv1}. While originally designed for resource constrained settings, MobileNetV1 has been found to be highly efficient across platforms and has been influential in the design of computer vision models\cite{mobilenetv2, efficientnet, mobilenetv3}.

MobileNetV1 is made up of alternating depthwise and 1$\times$1 convolutions. Each convolution is followed by batch normalization \cite{batch-normalization} and a ReLU non-linearity. MobileNetV1 defines a range of models with size controlled by a width multiplier. The 1$\times$1 convolutions in these models are responsible for the large majority of the FLOPs and can be computed as matrix multiplication if the input data is stored in CHW format.

\subsubsection{Experimental Setup}
We introduce sparsity into the 1$\times$1 convolutions of MobileNetV1 using magnitude pruning \cite{to-prune-or-not}. We prune all models to 90\% sparsity. We leave the first layer dense, as we found it to be bandwidth bound by the activation matrix and thus saw less benefits from weight sparsity. We trained our baseline models on the ImageNet \cite{imagenet} dataset with 32 accelerators for 100 epochs. As we target efficient inference in the regime where inference costs outweigh training costs, we increase training time for our sparse models by 10$\times$ which helps the sparse models converge while being pruned.

At inference time, batch normalization can be fused into the preceding linear operation. We do this for all depthwise and 1$\times$1 convolutions. \rev{For depthwise convolution, we wrote kernels that support fused bias and ReLU operations}. We similarly fuse bias and ReLU into our sparse 1$\times$1 convolutions. For the 1$\times$1 convolutions in our dense baselines we use Nvidia cuBLAS, which is backed by highly-tuned assembly kernels. We additionally wrote a fused bias + ReLU kernel, which we use following these linear operations. For our sparse models, we use an oracle kernel selector for four 1$\times$1 convolutions where our heuristic was sub-optimal. For each model, we benchmark inference in single-precision on an Nvidia V100 GPU with a batch size of 1 image, as is common in online inference applications like self-driving cars.

\subsubsection{Results \& Analysis}

\begin{figure}[t!]
  \includegraphics[width=\columnwidth]{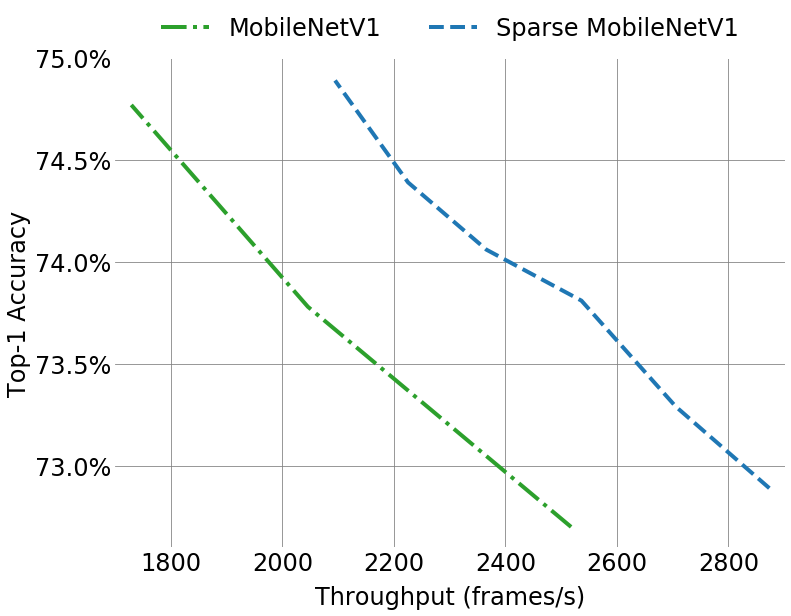}
  \vspace{-6mm}
  \caption{\textbf{MobileNetV1 accuracy-runtime tradeoff curves.} Sparse modes are more efficient, achieving speedups of \rev{21-24\%} for a given accuracy across all model sizes, or equivalently, \rev{\textasciitilde1.1\%} higher accuracy for the same throughput.
  }
\label{sparse_mbv1}
\end{figure}

The results of our benchmarks are shown in Figure \ref{sparse_mbv1} and in Table \ref{sparse_mbv1_table}. Across the board, our sparse models offer speedups of \rev{21-24\%} for a given accuracy, or equivalently, \rev{1.1\%} higher accuracy for the same throughput.

Because of the accuracy loss from pruning, our sparse models are wider than the dense ones they match accuracy with. Increased width also increases the cost of the non-sparse operations relative to those in the dense baseline. Better pruning algorithms would help alleviate this and enable further speedups. Additionally, the depthwise convolutions become a significant bottleneck after the 1$\times$1 convolutions are pruned. Tuning these kernels would yield further gains for our sparse models relative to their dense counterparts.

\begin{table}[t]
\caption{\textbf{Sparse Transformer Results}}
\label{sparse_transformer}
\centering

\begin{tabular}{cc|cc}
\hline
\multicolumn{2}{c|}{\textbf{Model}} & Transformer & Sparse Transformer \\ \hline
\multicolumn{2}{c|}{\textbf{Bits Per Dimension}} & 3.76 & 3.77 \\ \hline
\multirow{2}{*}{\textbf{V100}} & \textbf{Throughput (tokens/s)} & 32,477 & \rev{67,857} \\
& \textbf{Memory Usage (GB)}  & 9.88 & 0.77 \\ \hline
\multirow{2}{*}{\textbf{1080}} & \textbf{Throughput (tokens/s)} & out-of-memory & \rev{32,039} \\
& \textbf{Memory Usage (GB)} & out-of-memory & 0.88
\end{tabular}
\end{table}

\section{\rev{Related Work}}


\rev{\cite{fast-spmm} and \cite{magma-spmm} propose efficient SpMM kernels based on alternative sparse matrix formats designed for GPUs \cite{ellpack-r, sell-p}. \cite{spmm-design-principles} discuss the design of high-performance SpMM on GPUs. We compare to their approach for computing SpMM in Section \ref{kernel_benchmarks} and reference their taxonomy for SpMM design throughout the text. \cite{adaptive-sparse-tiling} propose an adaptive tiling technique, where CSR matrices are partitioned into sets of rows. Within each set, the columns are re-ordered such that columns with more nonzeros are grouped. These "heavy" groups are processed together and exploit tiled execution to enable more reuse of operands. The remaining columns are processed with a standard row-splitting scheme. We benchmark and discuss limitations of this approach in Section \ref{kernel_benchmarks}.}

\rev{\cite{direct-sparse-conv-intel} implement an efficient direct sparse convolution for CPUs and demonstrate performance gains relative to dense baselines. \cite{sparse-winograd-intel} develop a technique for inducing sparsity in Winograd convolutions \cite{winograd-convolution} and design and efficient implementation for CPUs. \cite{fast-sparse-convnets} design efficient SpMM kernels for CPUs and demonstrate significant performance improvements for highly efficient neural networks on mobile processors.}

\rev{\cite{blocksparse-gpu-kernels} and \cite{balanced-sparsity} enforce different forms of structure on sparse matrices to enable efficient mapping to GPUs. \cite{blocksparse-gpu-kernels} develop efficient GPU kernels for block-sparse matrices and apply them to neural networks on a range of different tasks. \cite{balanced-sparsity} propose fixing the number of nonzeros in small regions of the sparse matrix to balance between performance and accuracy loss from enforcing structure on the topology of nonzeros.}

\section{\rev{Discussion \&} Conclusion}
\rev{In addition to the kernels we discuss, training DNNs requires the computation $A^TB {\Rightarrow} C$, where $A^T$ is the transpose of a sparse matrix. It's difficult to fuse the transpose into the SpMM for CSR matrices. However, for DNN training it's possible to cache the row offsets and column indices for $A^T$ when the sparse matrix topology is updated and perform the transpose as an} \verb!argsort! \rev{of the matrix values. Alternative sparse matrix formats are an interesting direction to enable transposed and non-transposed computation \cite{compressed-sparse-blocks, hicoo}}

\begin{table}[t]
\caption{\textbf{Sparse MobileNetV1 Results}}
\label{sparse_mbv1_table}
\centering

\begin{tabular}{cc|cc}
\hline
\multicolumn{2}{c|}{\textbf{Model Width}} & \textbf{Accuracy} & \textbf{Throughput (frames/s)} \\ \hline
\multirow{3}{*}{\textbf{Dense}}   & 1    & 72.7\% & \rev{2,518} \\
                                  & 1.2  & 73.8\% & \rev{2,046} \\
                                  & 1.4  & 74.8\% & \rev{1,729} \\ \hline
\multirow{6}{*}{\textbf{Sparse}}  & 1.3  & 72.9\% & \rev{2,874} \\
                                  & 1.4  & 73.3\% & \rev{2,706} \\
                                  & 1.5  & 73.8\% & \rev{2,537} \\
                                  & 1.6  & 74.1\% & \rev{2,366} \\
                                  & 1.7  & 74.4\% & \rev{2,226} \\
                                  & 1.8  & 74.9\% & \rev{2,095}                         
\end{tabular}
\end{table}

\rev{On large problems, the performance of our kernels is limited by shared memory bandwidth. One direction for alleviating this bottleneck is to exploit reuse of values loaded from the right input across multiple rows of the left input matrix.}

\rev{While our kernels are highly efficient, they are not able to take advantage of dedicated matrix-multiply hardware. For unstructured sparsity, it's possible that unpacking sparse tiles in shared memory could enable the use of these operations. New advances in hardware are likely to enable this further \cite{ampere-whitepaper}. Despite model quality loss, it remains possible to exploit this hardware with vector and block sparsity \cite{blocksparse-rnn, blocksparse-gpu-kernels, fast-sparse-convnets}.}

In this work, we demonstrate that the sparse matrices found in deep neural networks exhibit favorable properties that can be exploited to accelerate computation. Based on this insight, we design high-performance SpMM and SDDMM kernels targeted specifically at deep learning applications. Using our kernels, we demonstrate sparse Transformer and MobileNet models that achieve 1.2--2.1$\times$ speedups and up to 12.8$\times$ memory savings without sacrificing accuracy. We hope that our findings facilitate better support for sparsity in deep learning frameworks and more broadly enable widespread use of sparsity in deep learning.

\section*{Acknowledgements}
We are grateful to Rasmus Larsen and Deepak Narayanan for providing detailed feedback on drafts of this paper. We'd also like to thank Penporn Koanantakool for her help debugging our kernel benchmarks, Artem Belevich for his help with Bazel and Docker and the TensorFlow team for answering many questions. Finally, we'd like to thank Nvidia for their help with our cuSPARSE benchmarks.

This research was supported in part by affiliate members and other supporters of the Stanford DAWN project---Ant Financial, Facebook, Google, Infosys, NEC, and VMware---as well as Toyota Research Institute, Cisco, SAP, and the NSF under CAREER grant CNS-1651570. Trevor Gale is supported by the National Science Foundation Graduate Research Fellowship under Grant No. DGE-1656518. Any opinions, findings, and conclusions or recommendations expressed in this material are those of the authors and do not necessarily reflect the views of the National Science Foundation. Toyota Research Institute ("TRI") provided funds to assist the authors with their research but this article solely reflects the opinions and conclusions of its authors and not TRI or any other Toyota entity.

\bibliographystyle{IEEEtran}
\bibliography{IEEEabrv,main}

\end{document}